\documentclass[conference]{IEEEtran}
\usepackage{etoolbox}
\makeatletter
\patchcmd{\@makecaption}
  {\scshape}
  {}
  {}
  {}
\makeatother

\usepackage{times}

\usepackage[numbers]{natbib}
\usepackage{multicol}
\usepackage[bookmarks=true]{hyperref}
\usepackage{graphicx}
\usepackage{algorithm}
\usepackage{algorithmic}
\newcommand{\etal}{\textit{et al}. }
\newcommand{\ie}{\textit{i}.\textit{e}.}
\newcommand{\eg}{\textit{e}.\textit{g}.}

\begin{document}
\title{Full-Frame Scene Coordinate Regression for Image-Based Localization}

\author{\authorblockN{Xiaotian Li,
Juha Ylioinas and
Juho Kannala}
\authorblockA{Aalto University\\  firstname.lastname@aalto.fi}}

\maketitle

\begin{abstract}
Image-based localization, or camera relocalization, is a fundamental problem in computer vision and robotics, and it refers to estimating camera pose from an image. Recent state-of-the-art approaches use learning based methods, such as Random Forests (RFs) and Convolutional Neural Networks (CNNs), to regress for each pixel in the image its corresponding position in the scene's world coordinate frame, and solve the final pose via a RANSAC-based optimization scheme using the predicted correspondences. In this paper, instead of in a patch-based manner, we propose to perform the scene coordinate regression in a full-frame manner to make the computation efficient at test time and, more importantly, to add more global context to the regression process to improve the robustness. To do so, we adopt a fully convolutional encoder-decoder neural network architecture which accepts a whole image as input and produces scene coordinate predictions for all pixels in the image. However, using more global context is prone to overfitting. To alleviate this issue, we propose to use data augmentation to generate more data for training. In addition to the data augmentation in 2D image space, we also augment the data in 3D space.  We evaluate our approach on the publicly available 7-Scenes dataset, and experiments show that it has better scene coordinate predictions and achieves state-of-the-art results in localization with improved robustness on the hardest frames (\eg, frames with repeated structures).
\end{abstract}

\section{Introduction}
Image-based localization is an important problem in computer vision and
robotics. It addresses the problem of estimating the 6 DoF camera pose in an environment from an image. It is a key component of many computer vision applications such as navigating autonomous vehicles and mobile robotics \cite{FAB-MAP}, simultaneous localization and mapping (SLAM) \cite{Eade_2006}, and augmented reality \cite{Castle_2008,LynenSBHPS15}. 

Currently, there are plenty of image-based localization methods proposed in the literature. Many state-of-the-art approaches \cite{SattlerLK12,SattlerLK17} are based on local features, such as SIFT, ORB, or SURF \cite{Lowe04,RubleeRKB11,Bay06surf}, and efficient 2D-to-3D matching. Given a 3D scene model, where each 3D point is associated with the image features from which it was triangulated, localizing a new query image against the model is solved by first finding a large set of matches between 2D image features and 3D points in the model via descriptor matching, and then using RANSAC \cite{fischler_bolles_1981} to reject outlier matches and estimate the camera pose on inliers. Although these local feature based methods have been proven to be very accurate and robust in many situations, due to the limitations of the hand-crafted feature detector and descriptor, extremely large viewpoint changes, repetitive structures and textureless scenes often produce a large number of false matchings which will lead to localization failure. 

Recently, some promising neural network based localization approaches have been proposed. Approaches such as PoseNet \cite{kendall2015convolutional} formulate 6 DoF pose estimation as a regression problem, and thus no traditional feature extraction, feature description, or feature matching processes are required. It has been shown that these approaches overcome the limitations of the local feature based approaches, \ie, they are able to recover the camera pose in very challenging indoor environments where the traditional methods fail. However, their localization accuracy is still far below traditional approaches in other situations where local features perform well. Unlike PoseNet, approaches such as DSAC \cite{Brachmann_2017_CVPR} obtain the 6 DoF pose by a two-stage pipeline: first, regressing a 3D point position for each pixel in an image, and then determining the camera pose using RANSAC based on these correspondences as in the conventional localization pipeline. Results have shown that these methods are the current 
state-of-the-art on the standard 7-Scenes benchmark \cite{SCoRF}. 

\begin{figure*}
\begin{center}
\centering \includegraphics[height=4.4cm]{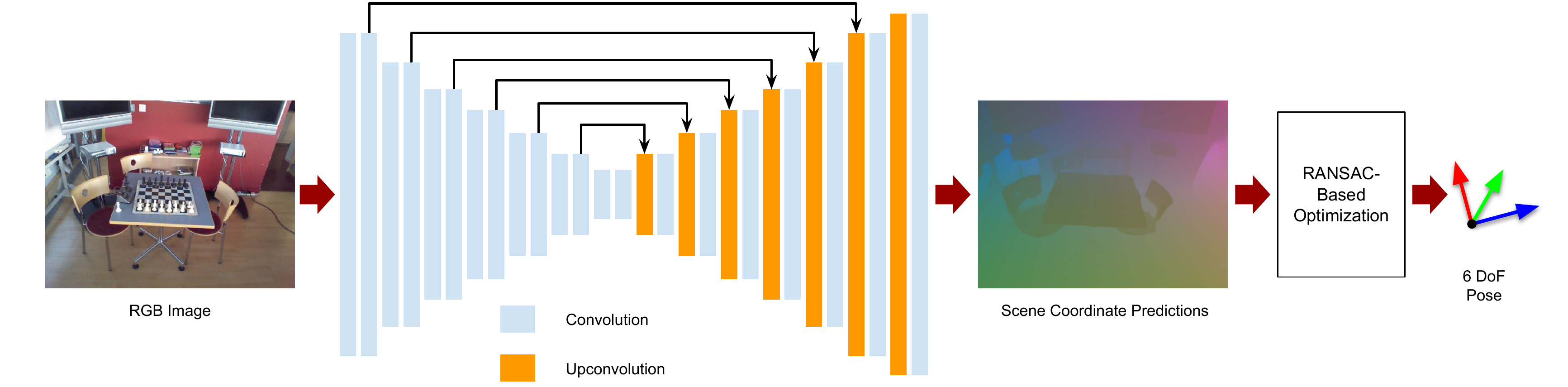}
\end{center}
\caption{Overview of our system. In this two-stage pipeline, the encoder-decoder full-frame Coordinate CNN accepts a whole RGB image as input and produces scene coordinate predictions for all pixels in the
image, and then the final pose is solved via a RANSAC-based optimization scheme using the predicted correspondences. \label{pipeline}} 
\end{figure*}

In this paper, we follow the two-stage RGB-only localization pipeline and present a full-frame Coordinate CNN for the scene coordinate regression in the first stage. This full-frame Coordinate CNN is trained to generate dense scene coordinate predictions from a given RGB image. Compared to the patch-based Coordinate CNN adopted in the DSAC pipeline \cite{Brachmann_2017_CVPR}, which is trained to generate a single scene coordinate prediction from an image patch centered around a pixel, our network considers more global information and can be evaluated efficiently at test time. Nevertheless, encoding more global context during regression is more prone to overfitting than using local patches since local patches are more invariant to viewpoint changes. Therefore, we propose to use data augmentation to mitigate the problem. We show with experiments on the widely used 7-Scenes dataset that our approach has improved  scene coordinate predictions and achieves state-of-the-art results in localization with strong robustness.

The contributions of this paper can be summarized as
follows:
\begin{itemize}
  \item We adopt a fully convolutional encoder-decoder network for the scene coordinate regression in the two-stage image-based localization pipeline to encode more global information.
  \item We do data augmentation in both 2D and 3D space for the scene coordinate regression, and we show that data augmentation is necessary for training the full-frame Coordinate CNN to achieve good localization performance.
  \item Our approach achieves state-of-the-art results and shows improved robustness on the hardest frames.
\end{itemize}

\section{Related Work}
Solving the image-based localization problem consists of estimating the 6 DoF camera pose of a query image in an arbitrary scene. This is also known as camera relocalization. More specifically, the localization system takes only one image as input, and then outputs camera pose according to a given representation of the scene. The representation of the scene depends on the approach used. It can be a database of images, a reconstructed 3D model, or a trained deep neural network. Here we only consider one image as input, although the image-based localization task can be extended by using a sequence of images as input. 

In the earliest stages, image-based localization was solved by treating it as a location recognition problem \cite{ZhangK06}. In these approaches, image retrieval techniques are often applied to determine the location of the query image by matching it to a database of images and finding the most similar images to it. The locations of the database images are known. Thus, the location of the query image can be estimated according to the known locations. Initially, these methods worked on databases consisting of tens of thousands of images and then they were improved to deal with more than a million images. However, the localization performance provided by these approaches is limited by the accuracy of the known locations of the database images \cite{SattlerLK11} and the density of the key-frames. 

To obtain better localization performance, more recent techniques \cite{IrscharaZFB09} are based on more detailed and structured information, \ie, they use a reconstructed 3D point cloud model, usually obtained
from Structure-from-Motion, to represent the scene. This is enabled by some powerful SfM approaches such as Bundler \cite{SnavelySS08}. It is even possible to construct huge models with millions of points \cite{StrechaPF10}. Thus, image-based localization systems are enabled to work on a city-scale level. Instead of matching the query image to the database images, these keypoint based methods solve the localization task by finding  correspondences between the query image and the 3D model. This is achieved by the use of the conventional hand-crafted features, \eg, SIFT \cite{Lowe04}. The features of the 3D points are computed during the 3D reconstruction, and the features of the query image are detected and extracted at test time. Therefore, the correspondence search is formulated as a descriptor matching problem. After establishing the 2D-3D correspondences, the camera pose is determined by 
a Perspective-n-Point \cite{KneipSS11} solver inside a RANSAC \cite{fischler_bolles_1981} loop.

It is obvious that the pose estimation step can only succeed if the quality of the established correspondences is good enough. When the 3D model is too big in size or too complex, an efficient and effective descriptor matching step is essential. There are several techniques in the literature to handle this problem, such as 2D-to-2D-to-3D matching \cite{IrscharaZFB09}, 3D-to-2D-matching \cite{LiSH10}, prioritized matching \cite{SattlerLK11} and co-visibility filtering \cite{SattlerLK12}. However, even if the matching process is effective and efficient enough, limitations of the hand-crated features will still cause these keypoint based localization systems to fail.

Recently, it has been shown that machine learning methods have great potential to tackle the problem of image-based localization. PoseNet \cite{kendall2015convolutional} demonstrates the feasibility of formulating 6 DoF pose estimation as a regression problem. The pose of a query image is directly regressed by a deep CNN with GoogLeNet \cite{SzegedyLJSRAEVR15} architecture pre-trained on large-scale image classification data. However, its performance is still below the state-of-the-art keypoint based methods and far from ideal for practical camera relocalization. In order to improve the accuracy of PoseNet, several variants have been proposed in recent papers. For example, LSTM-Pose \cite{Walch_2017_ICCV} makes use of LSTM units \cite{Hochreiter1997} on the CNN output to exploit the structured feature correlation. The LSTM units play the role of a structured dimensionality reduction on the feature vector and lead to drastic improvements in localization performance. Another variant, Hourglass-Pose \cite{MelekhovYKR17}, is based on hourglass architecture which consists of a chain of convolution and upconvolution layers followed by a regression part. The upconvolution layers are introduced to preserve the fine-grained information of the input image and this mechanism has been proven to be able to further improve the accuracy of image-based localization using CNN based architectures. Besides, it has been shown that the use of a novel loss function based on scene reprojection error can also significantly improve PoseNet's performance \cite{Kendall_2017_CVPR}.

Unlike PoseNet, the scene coordinate regression forests (SCoRF) approach \cite{SCoRF} adopts a regression forest \cite{Criminisi_2013_DFC} to generate 2D-3D matches from an RGB-D input image instead of directly regressing the camera pose. The final camera pose is then determined via a RANSAC-based solver. This whole pipeline is similar to the keypoint based localization approaches, but no traditional feature extraction, feature description, or feature matching processes are required. The original SCoRF pipeline is further improved by exploiting uncertainty in the model \cite{ValentinNSFIT15}. Training the random forest to predict multimodal distributions of scene coordinates results in increased pose accuracy. 

In order to localize RGB-only images as well, the original SCoRF is extended by utilizing the increased predictive power of an auto-context random forest \cite{BrachmannMKYGR16}. In the most recent DSAC paper \cite{Brachmann_2017_CVPR}, a neural network based SCoRF pipeline for RGB-only images is proposed.
In contrast to previous SCoRF pipelines, two CNNs are adopted for predicting scene coordinates and for scoring hypotheses respectively. Moreover, the conventional RANSAC is replaced by a new differentiable RANSAC, which enables the whole pipeline to be trained end-to-end. In \cite{Massiceti_icra}, Massiceti \etal further explore efficient versus non-efficient and RF- versus non-RF-derived NN architectures for camera localization and propose a new fully-differentiable robust averaging technique for regression ensembles.

\section{Method}
Our CNN based approach presented in this paper is  inspired by the state-of-the-art RGB-only DSAC pipeline \cite{Brachmann_2017_CVPR}, and it consists of two stages. In the first stage, a Coordinate CNN is adopted to generate 2D-3D correspondences from a given RGB-only image. In the second stage, a RANSAC-based scheme is performed to determine the final pose estimate. The overall pipeline is illustrated in Figure \ref{pipeline}.

The Coordinate CNN used in the original DSAC pipeline \cite{Brachmann_2017_CVPR} is trained and evaluated in a patch-based manner. Similar to the patch-based approaches for semantic segmentation, patches need to be sampled during both training and test time. The patch-based Coordinate CNN is trained using image patches of fixed size ($42\times42$) centered around pixels and the corresponding world coordinates of these pixels. At test time, patches of the same size are used as inputs of the Coordinate CNN to generate 2D-3D correspondences. However, obtaining a set of 2D-3D correspondences in such a way is very inefficient. In the DSAC pipeline \cite{Brachmann_2017_CVPR}, 1600 correspondences are needed for pose estimation. Thus 1600 patches should be extracted and fed into the deep neural network. This is obviously time-consuming and also requires much memory. Moreover, the fixed patch size limits the size of the receptive field, and thus limits the information that can be processed by the network. This makes it difficult for the Coordinate CNN to predict accurate 2D-3D correspondences when the scene exhibits ambiguities (\eg, repeated structures), since the global context cannot be used by the network. 

To overcome such limitations, we propose to use an encoder-decoder CNN which accepts a whole image as input and produces scene coordinate predictions for all pixels in the image. This is inspired by recently presented architectures for solving per-pixel tasks, such as semantic segmentation, depth estimation, and disparity estimation. Performing scene coordinate regression in this way, patches are not required to be sampled and the 2D-3D correspondences can be generated efficiently at test time. Since the receptive field of the encoder-decoder architecture covers almost the entire input image, more global context is added to the regression process and more overall information is considered. However, the global image appearance may change a lot due to only small viewpoint changes. That is, it is not as invariant to viewpoint changes as local appearance. Therefore, regressing the scene coordinates using more global information is prone to overfitting. We propose to use data augmentation to mitigate the problem. We call our network full-frame Coordinate CNN.

\subsection{Network Architecture}
Our full-frame Coordinate CNN is based on the architecture of DispNet \cite{MayerIHFCDB16}. It is fully convolutional \cite{LongSD15} such that dense per-pixel scene coordinate predictions can be generated from arbitrary-sized input images. We call the outputs scene coordinate images. The network consists of a contractive part and an expanding part. The input image is first spatially compressed via the contractive part and then
refined via the expanding part. Moreover, shortcut connections are added in between to overcome the data bottleneck. Unlike DispNet \cite{MayerIHFCDB16}, there is only one final output layer at the end of the network and no multi-scale side predictions are used. Instead of ReLU \cite{NairH10}, we use ELU \cite{ClevertUH15} for the nonlinearity between layers. The details of our network architecture are described in Table \ref{FFCCNN}. Layers start with upconv are upconvolutional layers and all other layers are convolutional. Also for each layer, the size of the kernel, the stride, the number of input channels, the number of output channels, and the input ($+$ is a concatenation) are given.
\begin{table}
\caption{Our network architecture.\label{FFCCNN}}
\centering
\begin{tabular}{|c|c|c|c|c|}
\hline
Name & Kernel & Str. & Ch I/O & Input \\ \hline
conv1a      & $7\times7$ & 2                         & 3/32                        & image                      \\ %\hline
conv1b      & $7\times7$ & 1                         & 32/32                        & conv1a                     \\ \hline
conv2a      & $5\times5$ & 2                         & 32/64                       & conv1b                     \\ %\hline
conv2b      & $5\times5$ & 1                         & 64/64                       & conv2a                     \\ \hline
conv3a      & $3\times3$ & 2                         & 64/128                      & conv2b                     \\ %\hline
conv3b      & $3\times3$ & 1                         & 128/128                     & conv3a                     \\ \hline
conv4a      & $3\times3$ & 2                         & 128/256                     & conv3b                     \\ %\hline
conv4b      & $3\times3$ & 1                         & 256/256                     & conv4a                     \\ \hline
conv5a      & $3\times3$ & 2                         & 256/512                     & conv4b                     \\ %\hline
conv5b      & $3\times3$ & 1                         & 512/512                     & conv5a                     \\ \hline
conv6a      & $3\times3$ & 2                         & 512/512                     & conv5b                     \\ %\hline
conv6b      & $3\times3$ & 1                         & 512/512                     & conv6a                     \\ \hline
conv7a      & $3\times3$ & 2                         & 512/512                     & conv6b                     \\ %\hline
conv7b      & $3\times3$ & 1                         & 512/512                     & conv7a                     \\ \hline
upconv6     & $3\times3$ & 2                         & 512/512                     & conv7b                     \\ %\hline
iconv6      & $3\times3$ & 1                         & 1024/512                     & upconv6+conv6b             \\ \hline
upconv5     & $3\times3$ & 2                         & 512/512                     & iconv6                     \\ %\hline
iconv5      & $3\times3$ & 1                         & 1024/512                     & upconv5+conv5b             \\ \hline
upconv4     & $3\times3$ & 2                         & 512/256                     & iconv5                     \\ %\hline
iconv4      & $3\times3$ & 1                         & 512/256                     & upconv4+conv4b             \\ \hline
upconv3     & $3\times3$ & 2                         & 256/128                     & iconv4                     \\ %\hline
iconv3      & $3\times3$ & 1                         & 256/128                     & upconv3+conv3b             \\ \hline
upconv2     & $3\times3$ & 2                         & 128/64                      & iconv3                     \\ %\hline
iconv2      & $3\times3$ & 1                         & 128/64                       & upconv2+conv2b             \\ \hline
upconv1     & $3\times3$ & 2                         & 64/32                       & iconv2                     \\ %\hline
iconv1      & $3\times3$ & 1                         & 64/32                       & upconv1+conv1b             \\ \hline
upconv0     & $3\times3$ & 2                         & 32/16                       & iconv1                     \\ %\hline
iconv0      & $3\times3$ & 1                         & 16/16                       & upconv0                    \\ \hline
coord\_pred & $3\times3$ & 1                         & 16/3                        & iconv0                     \\ \hline
\end{tabular}
\end{table}

\subsection{Training Loss}\label{sec:meth:loss}
Similar to the patch-based Coordinate CNN used in the original DSAC pipeline, we train our full-frame Coordinate CNN by minimizing the Euclidean distance between the scene coordinate ground truth and the prediction as given in Equation \ref{coordloss}. Unlike training in a patch-based manner where patches and single predictions are used and only patches centered at pixels with valid depth values are sampled for training, our network uses pairs of color image and scene coordinate image (dense scene coordinate values) as training samples where the ground truth scene coordinates can be missing for some pixels. Here we simply ignore the pixels without the ground truth scene coordinates and mask out their contributions to the final loss. More formally, loss for a training sample can be written as:
\begin{equation}
\label{coordloss}
loss = 	§  \sum_{i,j}\mathbf{M}_{ij}\| \hat{\mathbf{Y}}_{ij}-\mathbf{Y}_{ij}\| 
\end{equation}
where $\mathbf{Y}$ and $\hat{\mathbf{Y}}$ are ground truth scene coordinate image and scene coordinate image prediction respectively, $\mathbf{M}$ is a mask and $(i,j)$ is a 2D pixel coordinate. The mask $\mathbf{M}$ has the same resolution as the color image and the scene coordinate image. A pixel of $\mathbf{M}$ is set to 1 if the corresponding scene coordinate ground truth exists and 0 otherwise.

\subsection{Data Augmentation}\label{sec:meth:aug}
While our network has larger receptive field and can encode more global context information for better understanding of the scene, it is more prone to overfitting compared to a network trained and tested in a patch-based manner. This is because local patches are more invariant to viewpoint changes and we can usually sample sufficiently enough patches for training. However, in our case, we need more training images taken from a lot of different viewpoints to regularize the network, since global image appearance is more sensitive to viewpoint changes. Unfortunately, a common problem for image-based localization is that the training poses in the whole pose space is not complete enough to regularize such a network. Also, image-based localization is typically performed scene-wise, which means that only the images belonging to a particular scene can be used during training. It does not make sense to use data belonging to other scenes, and unlike for other computer vision tasks such as semantic segmentation, it is not straightforward to apply transfer learning technique for image-based localization. Hence, overfitting can be a serious problem in training our full-frame Coordinate CNN for solving image-based localization.

To alleviate the overfitting problem, we propose to use data augmentation to generate more data for training. A simple and common way of doing data augmentation in the context of solving per-pixel tasks is to apply 2D affine geometric transformations to the training images. In our case, we apply 2D transformations to both the input RGB images and the ground truth scene coordinate images. The transformations we perform include translation, rotation, and scaling.

However, these 2D affine transformations do not always generate correct input RGB images which are consistent with 3D scenes (although they are always consistent with their transformed ground truth scene coordinate images). Since the camera is in 3D space and  3D points are projected onto the image plane according to a camera model, if  camera pose is changed, the transformation between the old and new 2D coordinates of 3D points is much more complex than the combination of 2D transformations in the image plane. Therefore, we propose to augment data also in 3D space. For each training sample, since we already have the ground truth scene coordinate image, we can generate a local 3D point cloud from that. We then add a small random transformation to the ground truth pose to synthesize a new camera pose. Finally, using the new camera pose, the local 3D point cloud and the known camera model, we project the points to the new camera plane to generate a new RGB image and its ground truth scene coordinate image.

\begin{figure}
\begin{center}
\centering \includegraphics[height=1.8cm]{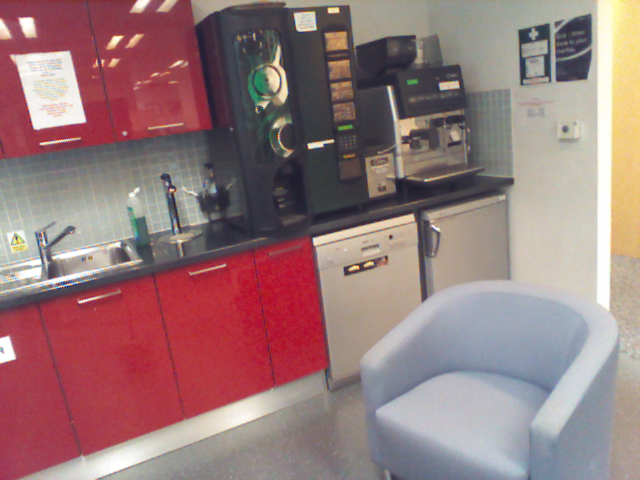}
~
\centering \includegraphics[height=1.8cm]{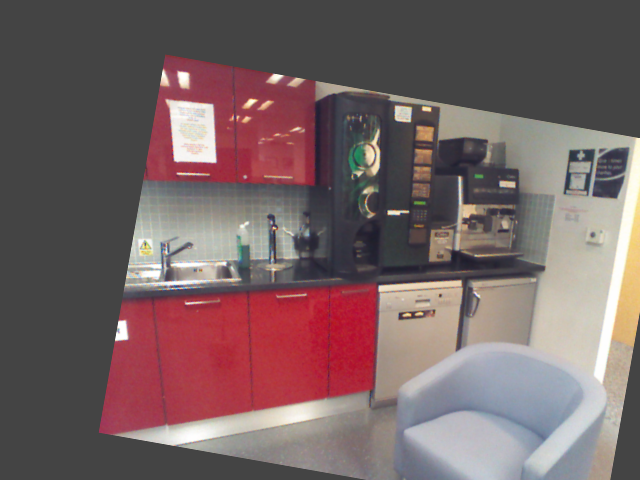}
~
\centering \includegraphics[height=1.8cm]{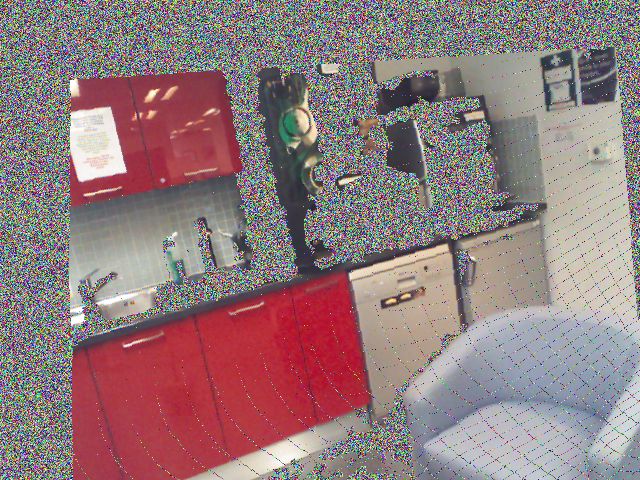}

\centering \includegraphics[height=1.8cm]{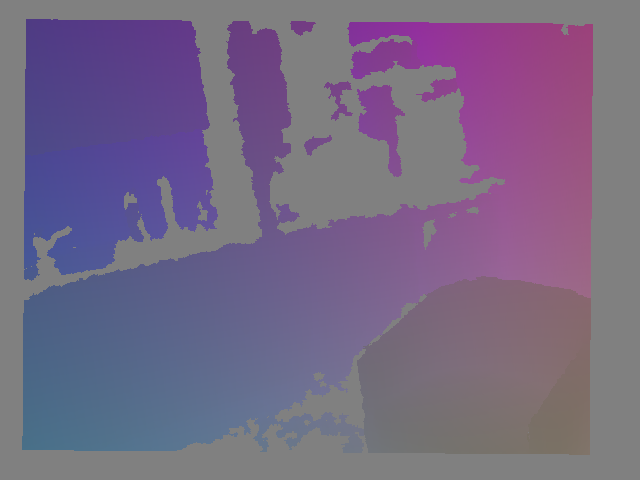}
~
\centering \includegraphics[height=1.8cm]{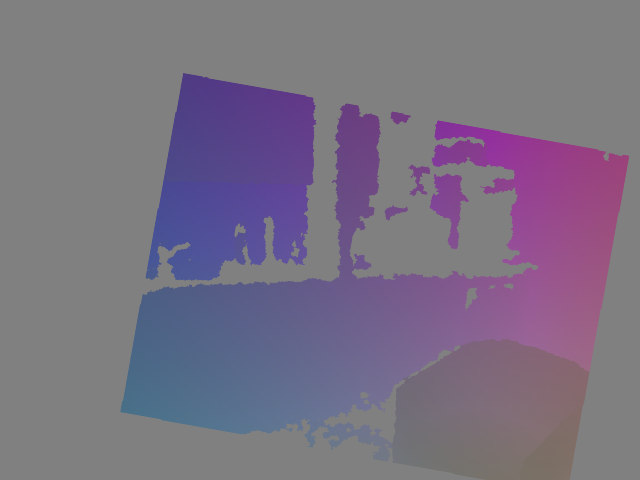}
~
\centering \includegraphics[height=1.8cm]{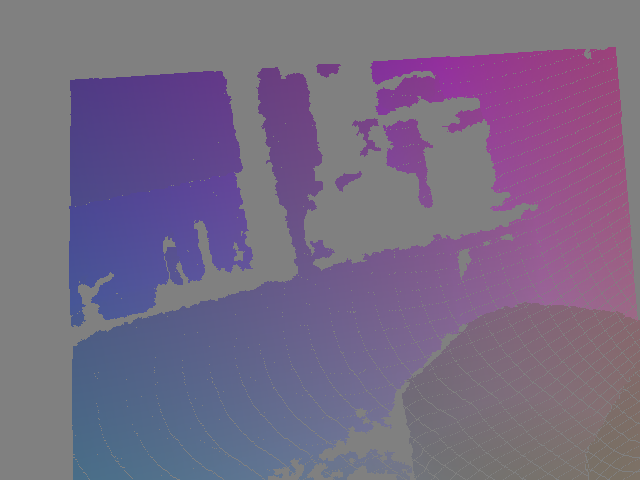}
\end{center}
\caption{Data augmentation. Left: original RGB and sence coordinate images. Middle: 2D transformation. Right: 3D transformation. \label{augex}}    
\end{figure} 

\subsection{Camera Pose Optimization}
In the second stage of the two-stage pipeline, a RANSAC algorithm is used in combination with a PnP algorithm to determine the final pose estimation from the 2D-3D correspondences generated in the first stage. Here the RANSAC pose optimization algorithm we use is similar to the one used in the DSAC pipeline \cite{Brachmann_2017_CVPR}. The only differences are that firstly we simply score the hypotheses by counting inliers instead of using a Score CNN \cite{Brachmann_2017_CVPR} to regress the scores of hypotheses from their reprojection error images, and secondly we use the conventional argmax hypothesis selection instead of the probabilistic selection. We believe the Score CNN used in \cite{Brachmann_2017_CVPR} is prone to overfitting.

The detailed steps of the RANSAC-based pose optimization algorithm used in this paper are described in Algorithm \ref{codeRANSAC}. Here, $score(H_i)$ is the number of inliers, \ie, the number of correspondences with reprojection errors less than a threshold $\tau$, and the final pose $H_f$ is the one with highest inlier count.  Following \cite{Brachmann_2017_CVPR}, $40\times40$ correspondences are sampled ($N=1600$), 256 hypotheses are sampled per image ($K=256$), the inlier threshold $\tau$ is set to 10 pixels, 8 refinement steps are performed ($R=8$), in each refinement step at most 100 inliers are chosen ($P=100$), and the refinement is stopped in case less than 50 inliers have been found ($Q=50$).

\begin{algorithm}
\caption{Pseudocode for RANSAC optimization}
\label{codeRANSAC}
\begin{algorithmic} 
\STATE  01: evaluate the Coordinate CNN to obtain $N$ correspondences
\STATE  02: $i\leftarrow 0$
\STATE  03: \textbf{while} $i<K$ \textbf{do} 
\STATE  04: $\ \ \ \ $sample $4$ 2D-3D correspondences
\STATE  05: $\ \ \ \ $generate a pose hypothesis $h$ using PnP
\STATE  06: $\ \ \ \ $\textbf{if} all the 4 correspondences are inliers of $h$ \textbf{then} 
\STATE  07: $\ \ \ \ \ \ \ \ H_i\leftarrow h$
\STATE 08: $\ \ \ \ \ \ \ \ S_i\leftarrow score(H_i)$
\STATE 09: $\ \ \ \ \ \ \ \ i\leftarrow i + 1$
\STATE 10: select the final pose $H_f$ according to the scores
\STATE 11: $j\leftarrow 0$
\STATE 12: \textbf{while} $j<R$ \textbf{do} 
\STATE 13: $\ \ \ \ $sample at most $P$ inliers 
\STATE  14: $\ \ \ \ $\textbf{if} less than $Q$ inliers are found \textbf{then} 
\STATE  15: $\ \ \ \ \ \ \ \ $\textbf{break} 
\STATE 16: $\ \ \ \ $recalculate the pose $H_f$ from the inliers
\STATE 17: \textbf{return} pose $H_f$
\end{algorithmic}
\end{algorithm}

\section{Experiments}
In this section, we evaluate the performance of our system, and compare it with other state-of-the-art methods. We use the 7-Scenes dataset \cite{SCoRF} for benchmarking.

\subsection{Evaluation Dataset}
For the experiments, we use the 7-Scenes dataset \cite{SCoRF} provided by Microsoft Research. The 7-Scenes dataset is a widely used RGB-D dataset which consists of seven different indoor environments. The RGB-D images are captured using a handheld Kinect camera at $640\times480$ resolution and associated with 6 DoF ground truth camera poses obtained via the KinectFusion system. A dense 3D model is also available for each scene. Each scene contains multiple sequences of tracked RGB-D camera frames and these sequences are split into training and testing data. The dataset exhibits several challenges, namely motion blur, illumination changes, textureless surfaces, repeated structures (\eg, in the Stairs dataset), reflections (\eg, in the Redkitchen dataset), and sensor noise. Following \cite{Brachmann_2017_CVPR}, we use the depth images to generate the ground truth scene coordinates. At test  time, we only use RGB images to estimate poses. 

\subsection{Implementation Details}
We train our network
from scratch for 800 epochs with a batch size of 16 (\ie, 200k updates for Chess) using the Adam optimizer where $\beta_1=0.9$, $\beta_2=0.999$, and $\epsilon = 10^{-8}$. The loss is computed as described in Section \ref{sec:meth:loss}. The initial learning rate is set to $0.0001$ and is halved every 200 epochs until the end.  

As mentioned in Section \ref{sec:meth:aug}, we perform data augmentation online during network training. We perform the 2D affine transformation with a 40\% chance, perform the 3D transformation with a 50\% chance, and use the original image with a 10\% chance. For the 2D transformation, we uniformly sample translation from the range
[$-20\%$, $20\%$] of the image width and height for $x$ and $y$ respectively, sample rotation from [$-45^{\circ}$, $45^{\circ}$] and sample scaling from [0.7, 1.5]. All pixels without valid RGB values are padded with the same random value. For the 3D transformation, we uniformly sample the rotational axis and the rotational angle is uniformly sampled from [$0^{\circ}$, $60^{\circ}$]. The direction of the translation vector is again uniformly sampled, and its magnitude in mm is sampled from  [$0$, $200$]. All pixels without valid RGB values are padded with different random values. Figure \ref{augex} shows an example of the data augmentation.  

At test time, although our full-frame Coordinate CNN can generate $640\times480$ scene coordinate predictions, we only sample $40\times40$ of the predictions for the pose estimation stage to make it consistent with the DSAC pipeline \cite{Brachmann_2017_CVPR}.

\subsection{Results}
We report both median error of camera pose estimations and accuracy measured as the percentage of query images for which the camera pose error is below $5^{\circ}$ and 5cm. The results of our method on the 7-Scenes dataset together with two current state-of-the-art methods are shown in Table \ref{re_all}. Following \cite{Brachmann_2017_CVPR}, Complete denotes the combined set of frames (17000) of all scenes. We also plot the cumulative histograms of localization errors for a more detailed comparison and report the results of \cite{SattlerLK12} and DSAC* (\ie, DSAC \cite{Brachmann_2017_CVPR} patch-based scene coordinate regression + conventional RANSAC). Note that for the DSAC pipeline \cite{Brachmann_2017_CVPR}, only the accuracy is reported in the original paper, and we obtain the additional results using the publicly available source code and trained models provided by the authors. For \cite{SattlerLK12}, we use Colmap \cite{SchonbergerF16} to construct models from scratch and register the models against the ground truth poses of the training images, and again we use the publicly available implementation of \cite{SattlerLK12}.

\begin{table*}
\caption{Results of our approach on the 7-Scenes dataset. Our approach
outperforms two state-of-the-art methods and has significantly improved performance on Stairs. Training
the full-frame Coordinate CNN without our proposed data augmentation
dramatically degenerates the localization performance. \label{re_all}}
\begin{center}
\begin{tabular}{c|c|c|c|c|c|c|c|c|}
\cline{2-9}
                                  & \multicolumn{4}{c|}{$5^{\circ}$, 5cm}                                                                                                                                                  & \multicolumn{4}{c|}{Median Error}                                                                                                                                                                   \\ \hline
\multicolumn{1}{|c|}{Scene}       & \begin{tabular}[c]{@{}c@{}}Brachmann\\ \etal \cite{BrachmannMKYGR16}\end{tabular} & \begin{tabular}[c]{@{}c@{}}Brachmann\\ \etal \cite{Brachmann_2017_CVPR}\end{tabular} & \begin{tabular}[c]{@{}c@{}}Ours\\ noaug\end{tabular} & Ours   & \begin{tabular}[c]{@{}c@{}}Brachmann\\ \etal \cite{BrachmannMKYGR16}\end{tabular} & \begin{tabular}[c]{@{}c@{}}Brachmann\\ \etal \cite{Brachmann_2017_CVPR}\end{tabular} & \begin{tabular}[c]{@{}c@{}}Ours\\ noaug\end{tabular} & Ours                 \\ \hline
\multicolumn{1}{|c|}{Chess}       & 94.9\%                                                    & 94.6\%                                                    & 43.2\%                                               & 88.5\% & 1.5cm, $1.3^{\circ}$                                      & 2.1cm, $0.7^{\circ}$                                      & 5.9cm, $1.9^{\circ}$                                 & 2.4cm, $0.8^{\circ}$ \\ 
\multicolumn{1}{|c|}{Fire}        & 73.5\%                                                    & 74.3\%                                                    & 10.7\%                                               & 62.3\% & 3.0cm, $1.4^{\circ}$                                      & 2.8cm, $1.0^{\circ}$                                      & 13.2cm, $4.5^{\circ}$                                & 3.7cm, $1.4^{\circ}$ \\ 
\multicolumn{1}{|c|}{Heads}       & 48.1\%                                                    & 71.7\%                                                    & 14.1\%                                               & 75.1\% & 5.9cm, $3.4^{\circ}$                                      & 2.0cm, $1.3^{\circ}$                                      & 12.6cm, $9.4^{\circ}$                                & 2.4cm, $1.7^{\circ}$ \\ 
\multicolumn{1}{|c|}{Office}      & 53.2\%                                                    & 71.2\%                                                    & 30.6\%                                               & 73.1\% & 4.7cm, $1.7^{\circ}$                                      & 3.4cm, $1.0^{\circ}$                                      & 7.0cm, $2.0^{\circ}$                                 & 3.5cm, $1.0^{\circ}$ \\ 
\multicolumn{1}{|c|}{Pumpkin}     & 54.5\%                                                    & 53.6\%                                                    & 18.5\%                                               & 51.4\% & 4.3cm, $2.1^{\circ}$                                      & 4.7cm, $1.3^{\circ}$                                      & 10.5cm, $2.7^{\circ}$                                & 4.9cm, $1.3^{\circ}$ \\ 
\multicolumn{1}{|c|}{Red Kitchen} & 42.2\%                                                    & 51.2\%                                                    & 38.8\%                                               & 60.4\% & 5.8cm, $2.2^{\circ}$                                      & 5.0cm, $1.5^{\circ}$                                      & 6.1cm, $1.7^{\circ}$                                 & 4.2cm, $1.2^{\circ}$ \\ 
\multicolumn{1}{|c|}{Stairs}      & 20.1\%                                                    & 4.5\%                                                     & 0.2\%                                                & 29.5\% & 17.4cm, $7.0^{\circ}$                                     & 1.9m, $49.4^{\circ}$                                      & 43.4cm, $10.1^{\circ}$                               & 7.9cm, $2.1^{\circ}$ \\ \hline
\multicolumn{1}{|c|}{Average}     & 55.2\%                                                    & 60.1\%                                                    & 22.3\%                                               & 62.9\% & 6.1cm, $2.7^{\circ}$                                      & 30cm, $8.0^{\circ}$                                       & 14.1cm, $4.6^{\circ}$                                & 4.1cm, $1.4^{\circ}$ \\ \hline
\multicolumn{1}{|c|}{Complete}    & 55.2\%                                                    & 62.5\%                                                    & 28.0\%                                               & 64.9\% & -                                                         & 3.9cm, $1.6^{\circ}$                                      & 8.0cm, $2.3^{\circ}$                                 & 3.8cm, $1.2^{\circ}$ \\ \hline
\end{tabular}
\end{center}
\end{table*}

As we can see, the overall performance of DSAC* is better than the original DSAC \cite{Brachmann_2017_CVPR}. While achieving almost the same results on the easy frames, DSAC* has superior performance on the harder ones, \ie, DSAC* provides better 0.95 quantiles, especially for scenes that have fewer training images (Fire, Heads). More importantly, DSAC* is able to produce reasonable median localization error for Stairs, though the accuracy
on the hardest frames of Stairs is still terrible. The results of  DSAC* suggest that the use of the Score CNN \cite{Brachmann_2017_CVPR} could make the entire localization pipeline less robust.

\begin{figure*}
\begin{center}
\centering \includegraphics[height=14.5cm]{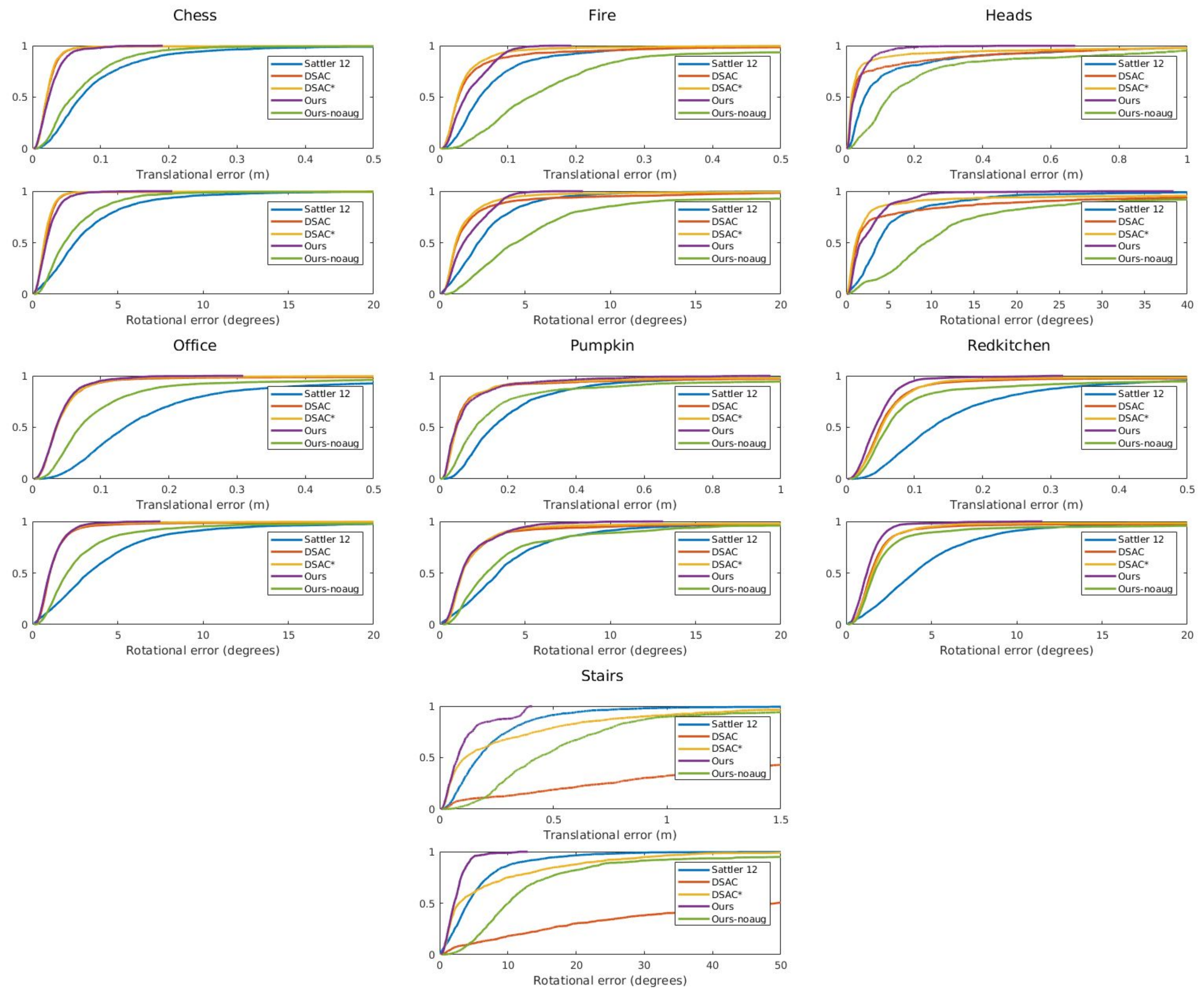}
\end{center}
\caption{Localization performance presented by cumulative histograms (normalized) of errors. Compared to other methods, our method is more robust, \ie, it has better performance on the hardest frames, since the curves always reach 1 very quickly. \label{loal_per_all}} 
\end{figure*}

According to the results, our proposed method outperforms the two state-of-the-art methods \cite{BrachmannMKYGR16,Brachmann_2017_CVPR} and the traditional keypoint based method \cite{SattlerLK12}, and its overall localization performance is more robust. To be specific, our method significantly improves the performance on the hardest frames. We find that while DSAC \cite{Brachmann_2017_CVPR} (as well as DSAC*) and \cite{SattlerLK12} often have extreme localization errors, \ie, rotational errors close to $180^{\circ}$ and translational errors larger than 2-5m, the maximum errors of our methods are always reasonable. As shown in Figure \ref{loal_per_all}, the \textit{Ours} curves always reach 1 very quickly compared to others. The results suggest that the global information could help the system to perform better in challenging scenarios. Remarkably, our method is able to provide significantly better localization performance for Stairs. Even the hardest frames can be localized with the errors less than 0.41m and $13^{\circ}$. It shows that the full-frame Coordinate CNN with enlarged receptive field can better cope with the repetitive structures, while the patch-based Coordinate CNN \cite{Brachmann_2017_CVPR} and the local keypoint based method \cite{SattlerLK12} are limited due to their local nature. Our method is also on-par with \cite{Massiceti_icra}. However, in \cite{Massiceti_icra}, only the overall accuracy is reported and this prevents us from carrying out a more detailed comparison. 

To show that our full-frame Coordinate CNN is more efficient at test time, we present the runtimes of the CNNs in Table \ref{runtimes}. We see that our full-frame Coordinate CNN is one order of magnitude faster than the patch-based one \cite{Brachmann_2017_CVPR}. And this comparison is even not fair since our full-frame Coordinate CNN produces $640\times480$ predictions while only $40\times40$ are generated by the patch-based one \cite{Brachmann_2017_CVPR}. In addition, our full-frame Coordinate CNN has a relatively small model size compared to the regression forest based model \cite{Massiceti_icra}. Our network has a model size of $\sim$120MB, while the most compact one from \cite{Massiceti_icra} has a model size of $\sim$250MB.

\begin{figure*}
\begin{center}
\centering \includegraphics[height=3.6cm]{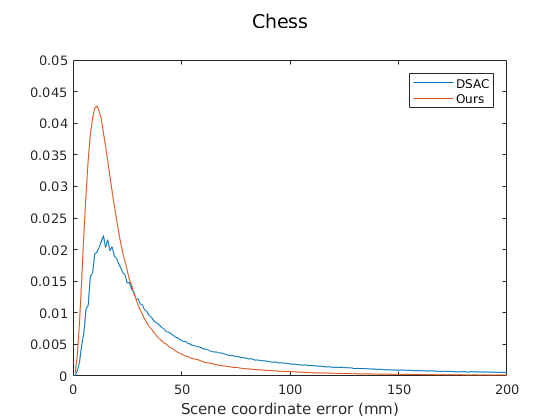}
~
\centering \includegraphics[height=3.6cm]{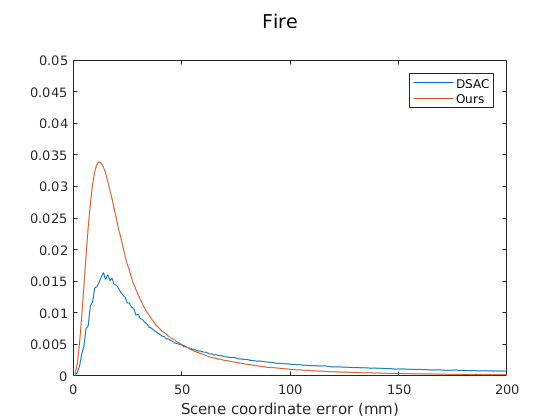}
~
\centering \includegraphics[height=3.6cm]{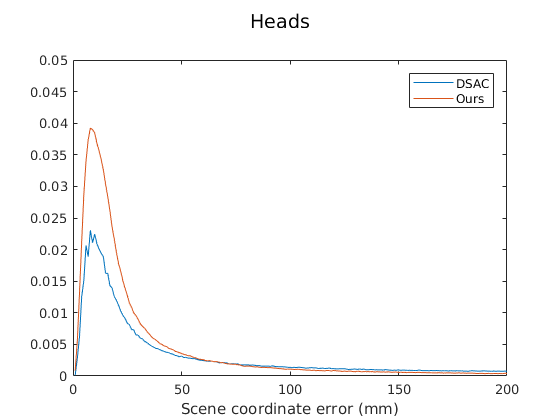}
~
\centering \includegraphics[height=3.6cm]{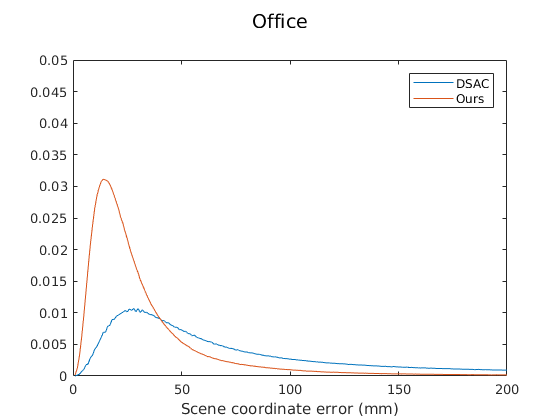}
~
\centering \includegraphics[height=3.6cm]{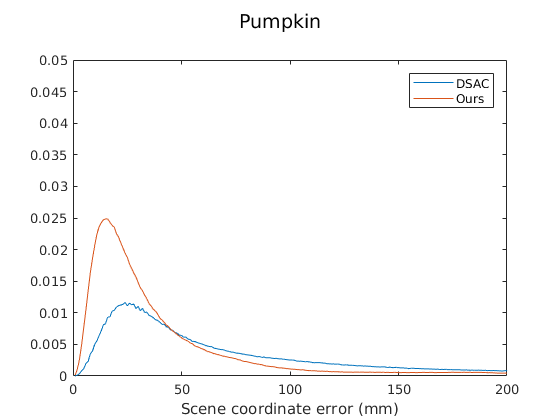}
~
\centering \includegraphics[height=3.6cm]{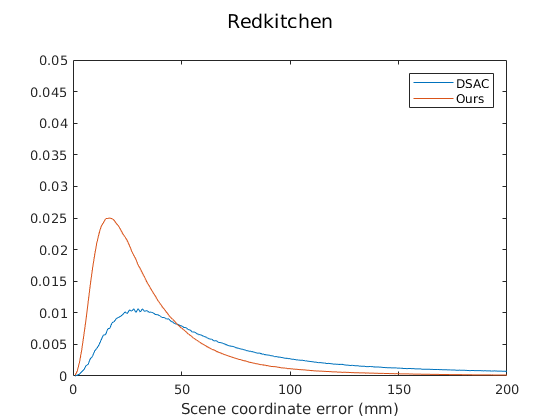}
~
\centering \includegraphics[height=3.6cm]{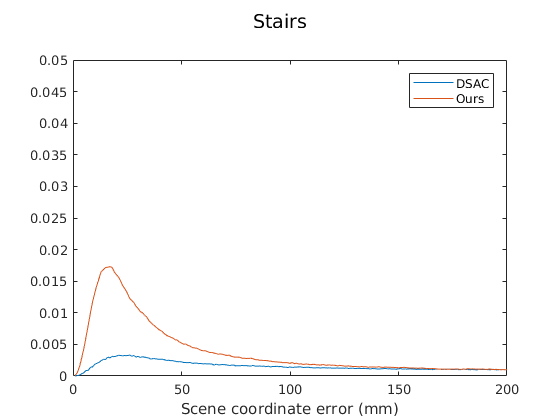}
\end{center}
\caption{Histograms (normalized) of scene coordinate errors. Our method provides better scene coordinate predictions. \label{coord_all}} 
\end{figure*}

Following \cite{Massiceti_icra}, in addition to the direct measure of localization performance, we present the accuracy of the intermediate scene coordinate predictions on the test images. In Table \ref{coord_err}, we report the percentage of scene coordinate inliers and the mean Euclidean distance between the inliers and their ground truth scene coordinate labels. A prediction is considered as an inlier if its Euclidean distance to its ground truth label is less than 10mm \cite{Massiceti_icra}. The normalized histograms of scene coordinate errors are illustrated in Figure \ref{coord_all}. As we can see, our full-frame Coordinate CNN is able to produce significantly better scene coordinate predictions. This shows why it is more robust than the patch-based one. However, this does not directly lead to equally better localization accuracy. This is because the RANSAC-based optimizer is highly robust and non-deterministic \cite{Massiceti_icra}.

To evaluate the effectiveness of data augmentation, we also report the performance of our method without the proposed data augmentation. According to the results, training the full-frame Coordinate CNN without data augmentation dramatically degenerates the localization performance on all the scenes. Without data augmentation, the full-frame Coordinate CNN is unable to generalize well to unseen images. This proves the importance of data augmentation which is used for regularizing our full-frame Coordinate CNN during training.

\begin{table}
\caption{The runtimes of the full-frame and patch-based Coordinate CNNs.\label{runtimes}}
\begin{center}
\begin{tabular}{|c|c|c|}
\hline
GPU       & Full-frame & Patch-based  \\ \hline
NVIDIA GeForce GT 750M        &$\sim$0.3s                          & $\sim$5s                              \\ %\hline
NVIDIA GeForce GTX 1080    & $\sim$0.02s               & $\sim$0.3s                              \\ \hline
\end{tabular}
\end{center}
\end{table}

\begin{table}
\caption{The percentages of the scene coordinate prediction inliers and the mean
errors of the inliers. Our predictions are clearly the most accurate.\label{coord_err}}
\begin{center}
\begin{tabular}{|c|c|c|}
\hline
DSAC \cite{Brachmann_2017_CVPR} & Best from \cite{Massiceti_icra} & Ours  \\ \hline
59.4\%, 41mm               & 45.3\%, 47mm                                    & 90.3\%, 28mm                                                       \\ \hline
\end{tabular}
\end{center}
\end{table}

\section{Conclusion}
In this paper, we present an image-based localization approach using fully convolutional encoder-decoder network to predict scene coordinates of RGB image pixels in a full-frame manner. The proposed full-frame Coordinate CNN can process more global information to better understand the scene. It takes a whole image as input and produces scene coordinate predictions for all pixels in the image and can be efficiently evaluated at test time. The proposed data augmentation is crucial for our approach to achieve good localization performance. The results show that our method provides more accurate scene coordinate predictions, achieves state-of-the-art localization performance and has improved robustness in challenging scenarios.

\bibliographystyle{plainnat}
\bibliography{bib}

\end{document}